\documentclass{article}
\usepackage{mathtools}
\usepackage[utf8]{inputenc} 
\usepackage[T1]{fontenc}    
\usepackage{hyperref}       
\usepackage{url}            
\usepackage{booktabs}       
\usepackage{amsfonts}       
\usepackage{nicefrac}       
\usepackage{microtype}      
\usepackage{xcolor}
\usepackage[super]{nth}
\usepackage{url}
\usepackage{amssymb}
\usepackage{algorithm} 
\usepackage{algorithmic}
\usepackage{subcaption}
\usepackage{multirow}
\usepackage{multicol}
\usepackage{amsmath,array}
\usepackage{natbib}
\usepackage{url}
\usepackage[margin=1.2in]{geometry}
\newcolumntype{A}{ >{$} r <{$} @{} >{${}} l <{$} }
\definecolor{Myblue}{rgb}{0,.3,.6}
\newcommand{\emc}[1]{{\textbf{\textit{\color{Myblue}#1}}}}

\DeclareMathOperator{\diag}{diag}

\DeclareMathOperator{\vecc}{vec}

\usepackage[T1]{fontenc}
\usepackage[utf8]{inputenc}
\usepackage[T1]{fontenc}

\usepackage{array}
\newcommand{\PreserveBackslash}[1]{\let\temp=\\#1\let\\=\temp}
\newcolumntype{C}[1]{>{\PreserveBackslash\centering}p{#1}}
\newcolumntype{R}[1]{>{\PreserveBackslash\raggedleft}p{#1}}
\newcolumntype{L}[1]{>{\PreserveBackslash\raggedright}p{#1}}
\title{
A Closed-Form Learned Pooling\\
for Deep Classification Networks
}
\date{}
\author{
  Vighnesh Birodkar
  \thanks{Work done as a Google AI Resident}\\
  \texttt{vighneshb@google.com}
  \and
  Hossein Mobahi\\
  \texttt{hmobahi@google.com}
  \and
  Dilip Krishnan\\
  \texttt{dilipkay@google.com}
    \and
  Samy Bengio\\
  \texttt{bengio@google.com}
}
\begin{document}

\maketitle

\begin{abstract}
In modern computer vision tasks, convolutional neural networks (CNNs) are indispensable for image classification tasks due to their efficiency and effectiveness. Part of their superiority compared to other architectures, comes from the fact that a single, local filter is shared across the entire image. However, there are scenarios where we may need to treat spatial locations in non-uniform manner. 
We see this in nature when considering how humans have evolved foveation to process different areas in their field of vision with varying levels of detail. 
In this paper we propose a way to enable CNNs to learn different  pooling weights for each pixel location. 
We do so by introducing an extended definition of a pooling operator. This operator can learn a strict super-set of what can be learned by average pooling or convolutions. It has the benefit of being shared across feature maps and can be encouraged to be local or diffuse depending on the data. We show that for fixed network weights, our pooling operator can be computed in closed-form by spectral decomposition of matrices associated with class separability. 
Through experiments, we show that this operator benefits generalization for ResNets and CNNs on the CIFAR-10, CIFAR-100 and SVHN datasets and improves robustness to geometric corruptions and perturbations on the CIFAR-10-C and CIFAR-10-P test sets.

\end{abstract}

\section{Introduction}

\label{sec:intro}
Convolutional Neural Networks (CNNs) have revolutionized the field of computer vision \citep{alexnet, resnetv2, maskrcnn}. Their success (compared to fully connected networks) is often attributed to their weight sharing in form of a convolution, which reduces the number of learnable parameters \citep{krizhevsky2012imagenet}. In addition, the ``shift invariance'' property of convolution has been believed to be crucial for improved generalization in vision tasks \citep{fukushima} (although some modifications may be required \citep{weiss, zhang}). Shift invariance, while crucial for handling translation in images, is a very limited form of real-world geometric transformations. For instance, convolutional representations are not invariant or equivariant to other basic transforms such as image rotation and scaling \citep{weiss}. 

There have been recent attempts to incorporate additional forms of invariances, such as rotation, reflection, and scaling.
\citep{sifre_mallat, bruna_mallat, esteves, kanazawa, worrall}. However, these methods engineer the invariance into networks; requiring a-priori identification of invariance types of interest. In this work, our goal is to achieve invariance using a data-driven approach to handle geometric transforms that may not fall into the categories mentioned above.

One way to achieve invariance or equivariance for transforms beyond translation  is non-uniform sub-sampling of the image (to be fed as input to a convolutional layer). For example, log-polar sampling of an image results in a new image, where rotation and scaling in the original image become equivalent to a translation in the resulted image \citep{polar}. This suggests that by \emc{adapting the pooling operator}, one can build representations that are better suited for variety of geometric transforms.

\begin{figure}
\centering
\begin{subfigure}[t]{0.40\textwidth}
\includegraphics[width=1\textwidth]{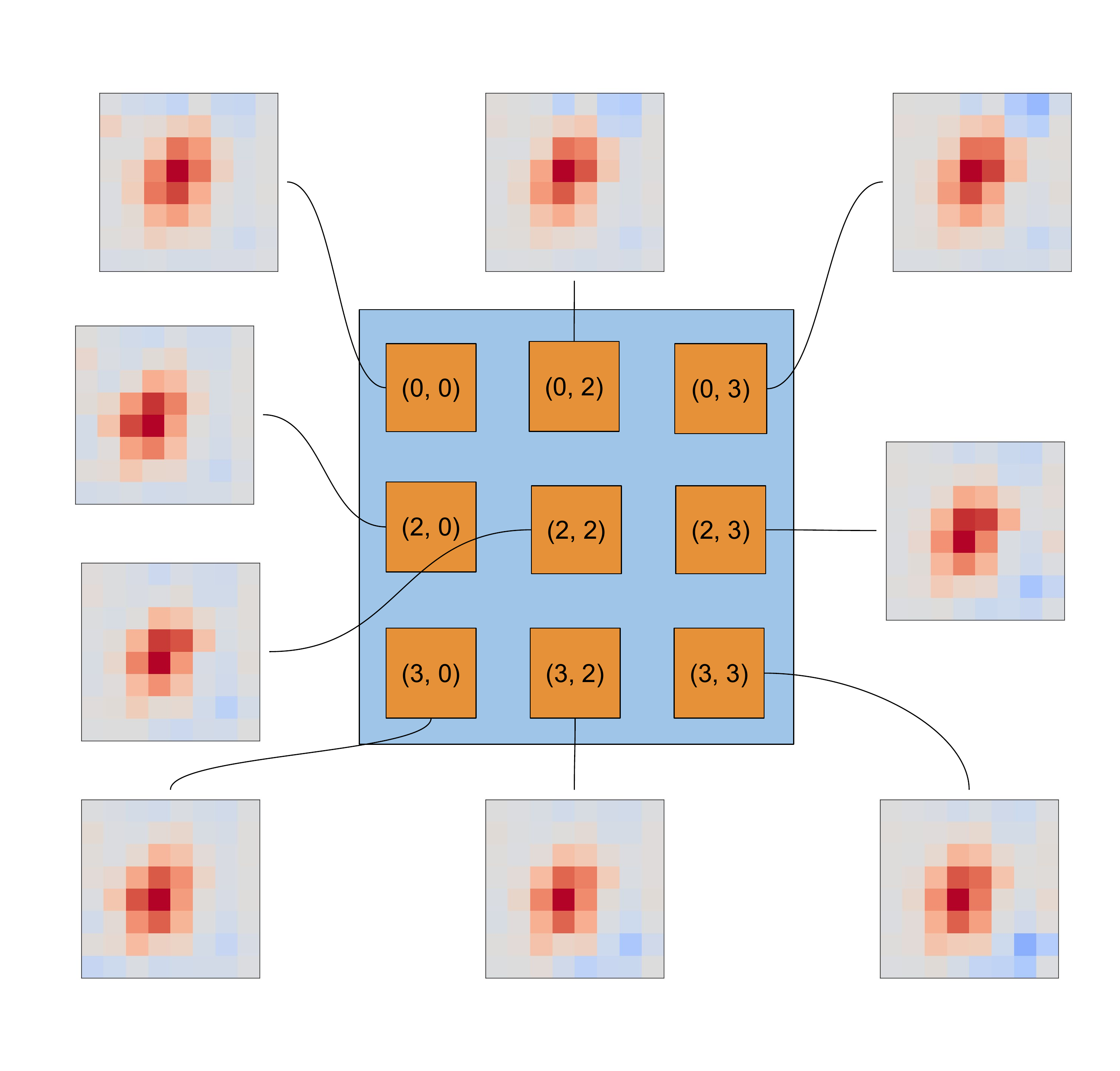}
\caption{Visualizing the pooling operator at 9 locations}
\label{fig:map_viz}
\end{subfigure}
\hspace{5pt}
\vline
\hspace{5pt}
\begin{subfigure}[t]{0.50\textwidth}
\includegraphics[width=1
\textwidth]{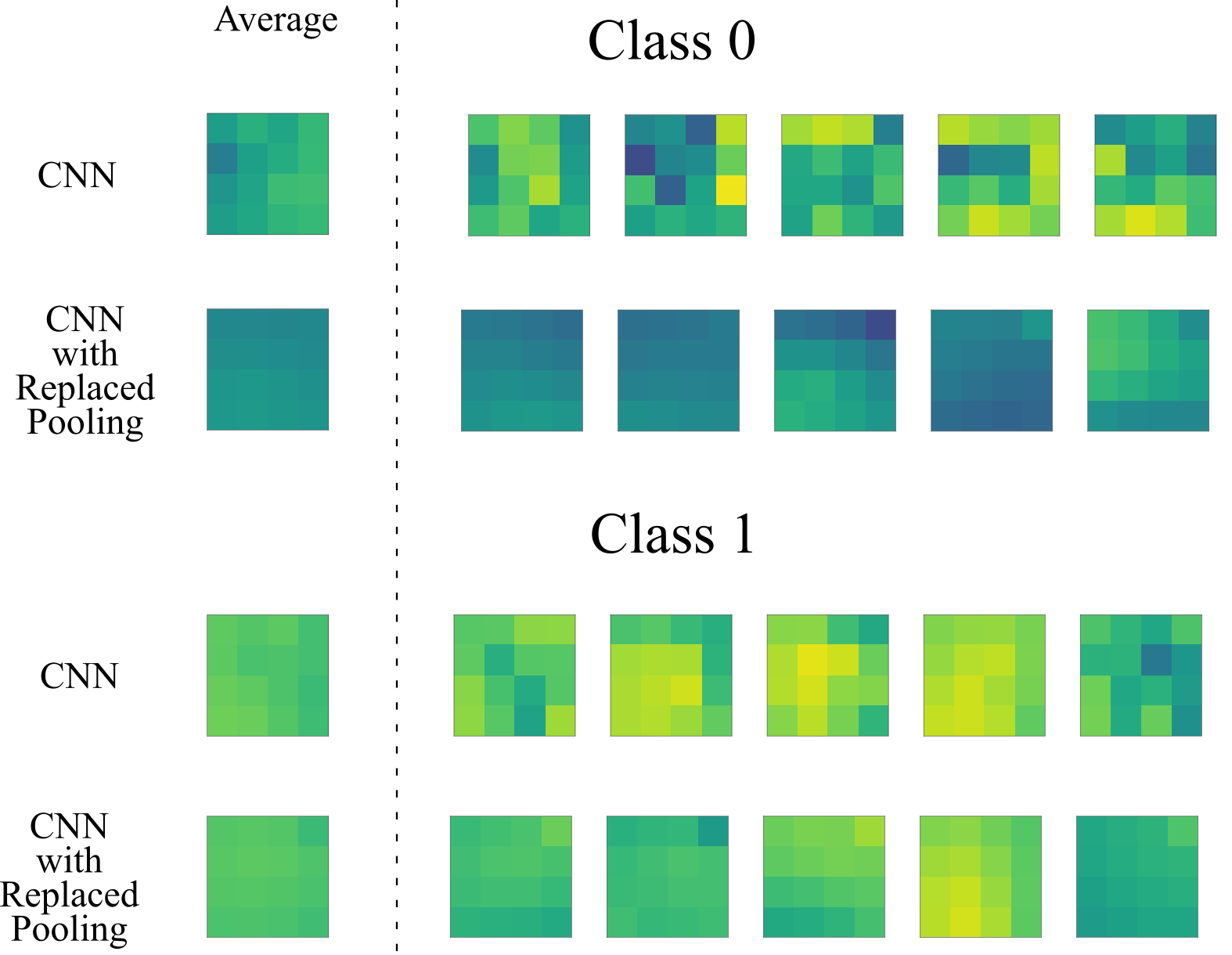}
\caption{Comparing the outputs of pooling operations.}
\label{fig:fmaps}
\end{subfigure}

\caption{ One the left we visualize the learned pooling map for a CNN model 
on the reduced SVHN dataset.
Each heat map corresponds to one of the 9 evenly spaced locations in the output 
feature map.
For each pixel in the output, we can see where the algorithm chooses to
put a positive(red) or negative(blue) weight over the input.
On the right, we compare the output of the default pooling used (strided convolution) against the output of our pooling operator.
We randomly select a channel and display the average value of
that channel per class along with 5 other values closest to the average.
We note that with our pooling operator, the per-class feature maps
are closer to the per-class average feature maps as compared to 
the regular CNN model.
}
\label{fig:vis_pooling}
\end{figure}

Non-uniform sampling is also the chosen scheme by nature; foveal vision implements a spatially-varying sampling similar to log-polar transform \citep{larson2009contributions}. Central and peripheral regions are sampled at different frequencies and both contribute to efficient and effective human vision. In addition, it is known that non-uniform sampling of image can facilitate image registration when geometric transforms are beyond translation \citep{hossein}. Our results with learned pooling operator confirms the advantage of spatially varying pooling. For example, Figure \ref{fig:map_viz} shows the response map of a learned pooling operator on SVHN dataset. The operator places more weights on the center pixel to take advantage of the fact that SVHN digits are mostly in the center. Contrast this with the operator learned for a CIFAR-100 model which places weights all across the spatial field (see Supplementary). The form of the learned pooling operator also affects the pooled feature maps. In Figure \ref{fig:fmaps}, the pooled feature maps are more clustered around the mean feature map of each class, compared to the feature maps produced by a regular CNN. This results in better separability of classes and better generalization as seen in Table \ref{tab:gen}.

In addition to adapting to the geometric transforms present in the data, and hence improving generalization, our learned pooling operator helps with robustness of the model. It has been observed that small geometric transforms in the image can result in prediction errors in existing deep models, which can be traced to the pooling operator \citep{weiss, zhang}; max pooling results in aliasing effects in the representation. While average pooling can prevent the aliasing issues (because it acts as a low-pass filter), the blurring causes loss of information and hence inferior classification performance to max pooling. However, by adapting the pooling operator to the data, in way that it can provide more class separability, relevant information are automatically picked up. In fact, our experiments show that the learned pooling can outperform the naive uniform down-sampling scheme that is used in most state-of-the-art models (strided pooling) and yet is robust to geometric perturbations on robustness benchmark datasets (CIFAR-10-C and CIFAR-10-P \citep{robustness}).

\section{Related Work}
Convolutional Neural Networks (CNNs) rely on pooling or sub-sampling to reduce the 
size of the hidden representation. This is known to have
important implications towards the kinds of invariances and generalization abilities of a network \citep{bias}. Earlier architectures have relied on average-pooling \citep{lenet} and max-pooling \citep{alexnet}, whereas modern ones learn parameters of pooling through strided convolutions \citep{resnet}.

The history of pooling in computer vision however goes past CNNs popularity. For instance, \citep{ask_locals} combines SIFT with pooling separately over learned clusters of features. \citep{fcn_pooling} learns pooling parameters of all spatial lower-level features through a fully connected network, whereas \citep{multi_scale} learns pooling separately at each scale using VLAD \citep{vlad}. \citep{dpm} defines distance transform pooling as part of deformable part models using a quadratic function of distance from the center and use a
latent SVM \citep{latent_svm} to learn it on top of a pre-trained CNN.
\citep{beyond_spatial} uses pooling at different scales in spatial dimensions and
also performs pooling on the color channels while aggregating using the max operator.

Pooling has also been used to aggregate varying input sizes into a fixed size representation.
\citep{bof} uses a fixed number of RBF neurons on top of a regular CNN to output a fixed size representation irrespective of the input image size. \citep{vid_bof} uses a specially designed pooling function for a multi-instance learning setting to output tags
for a video from tags predicted for each frame. \citep{learn_pool_vid} uses 
NetVLAD defined in \citep{net_vlad} and approximations of Bag of Words and Fischer vector encoding to aggregate features across time for learning video classification.

Recently, there have been some attempts to to learn parameters of local pooling operations of CNNs in an end-to-end fashion though gradient descent. \citep{leap} propose learning a local pooling operator, one per each channel and train a deep-neural network for classification.
\citep{dpp} try to preserve small details in the input while pooling and introduce two new parameters per input feature map to control which details
are preserved. \citep{gen_pool} experiment with learned and fixed 
combinations of average/max pooling and also suggest  organizing the outputs
of multiple local filters in the form of a binary tree to learn
the parameters of mixing them. 
Although these approaches learn pooling parameters from data,
the pooling operator is limited to spatially uniform; the same sampling scheme is used to pool each output pixel. As we discuss in Section \ref{sec:intro}, spatially
varying pooling is necessary to learn efficient and robust representation for
transformations other than translation.

\section{Method}
\subsection{Notation}

We first formalize the definition of linear pooling. Let $\mathcal{N}_k \triangleq \{1,2,3,\dots,k\}$. Given a spatial domain $\Omega \triangleq \mathcal{N}_I \times \mathcal{N}_J$ and set of intensity values $\Gamma$, a feature map of depth $C$ is a map $f:\Omega \times \mathcal{N}_C \rightarrow \Gamma$. We can represent a feature map $f$ in matrix form:
\begin{eqnarray}
{\boldsymbol{F}_c}_{\color{red} I \times J} &\triangleq& \begin{pmatrix} 
f(1,1,c) & f(1,2,c) & \dots & f(1,J,c) \\
f(2,1,c) & f(2,2,c) & \dots & f(2,J,c) \\
\vdots & \vdots & \vdots & \vdots \\
f(I,1,c) & f(I,2,c) & \dots & f(I,J,c)
\end{pmatrix} \\
{\boldsymbol{f}_c}_{\color{red} N \times 1} &\triangleq& \vecc(\boldsymbol{F}_c) \\
\boldsymbol{X}_{\color{red} N \times C} &\triangleq& \begin{pmatrix} 
\boldsymbol{f}_1 &|& \boldsymbol{f}_2 &|& \dots &|& \boldsymbol{f}_C
\end{pmatrix} \,.
\end{eqnarray}
where $N$ denotes the domain size $N \triangleq | \Omega| = I \times J$ and $\vecc$ converts a matrix into a column vector by concatenating the columns of the matrix.

We define \emc{linear pooling} as the operator $\boldsymbol{P}_{M \times N}$ which maps $\boldsymbol{X}_{\color{red} N \times C}$ into another feature map $\boldsymbol{Z}_{\color{red} M \times C}$ where $1 \leq M \leq N$. That is, the operator may shrink the input spatially but maintains the number of channels. Formally, the output of the linear operator is an element in $\omega \times \mathcal{N}_C$, where $\omega \triangleq \mathcal{N}_{I^\prime} \times \mathcal{N}_{J^\prime}$ and $M \triangleq |\omega| = I^\prime \times J^\prime$. Note that the operator is applied to each channel (column) of the input matrix to generate the corresponding channel (column) in the output matrix.
\begin{equation}
\label{eq:def_pooling}
\boldsymbol{Z}_{\color{red} M \times C} \,\triangleq\, \boldsymbol{P}_{\color{red} M \times N} \, \boldsymbol{X}_{\color{red} N \times C}
\end{equation}

\subsection{Formulation}

Obviously, \emc{average pooling} is seen as a special case for the linear operator $\boldsymbol{P}$ when the entries of the operator are set in a specific way. However, one may wonder if, within the space of all linear operators, there could be better choices than average pooling. Trivially, the answer is data dependent and hence we let the data itself discover the operator that suits the task in hand. In the classification setting, a good operator should help with classification. One possible way to quantify helping with classification is to improve separability of the classes, as explained below.

To simplify the formulation, we focus on finding each row of $\boldsymbol{P}$ separately. Let $\boldsymbol{z}_m$ be the $m$'th row of $\boldsymbol{Z}$, arranged as a column vector. Similarly, let $\boldsymbol{p}_m$ be the $m$'th row of $\boldsymbol{P}$, also arranged as a column vector. Then the pooling identity in (\ref{eq:def_pooling}) can be equivalently expressed using vectors as: 
\begin{equation}
{\boldsymbol{z}_m}_{\color{red} C \times 1} \,\triangleq\, {\boldsymbol{X}^T}_{\color{red} C \times N} \, {\boldsymbol{p}_m}_{\color{red} N \times 1} \quad,\quad m=1,\dots,M \,.
\end{equation}

To reduce mathematical clutter, we drop the index $m$ from $\boldsymbol{z}_m$ and $\boldsymbol{p}_m$. The reader should remember that the following result needs to be applied for each choice of $m=1,\dots,M$ separately. Hence with abuse of notation we proceed as:
\begin{equation}
\label{eq:def_z}
\boldsymbol{z}_{\color{red} C \times 1} \,\triangleq\, {\boldsymbol{X}^T}_{\color{red} C \times N} \, \boldsymbol{p}_{\color{red} N \times 1} \,.
\end{equation}

To define separability, we require a training set. Consider a set of $K$ feature maps $\boldsymbol{X}_1 ,\dots, \boldsymbol{X}_K$, whose elements are associated with a label $y_k \in \{1,\dots,Q\}$, with $Q$ being the number of classes in the dataset. We define the following total and per class average quantities:
\begin{equation}
\label{eq:def_avr}
\overline{\boldsymbol{z}}_{\color{red} C \times 1} \triangleq \frac{1}{K} \sum_k {\boldsymbol{z}_k}_{\color{red} C \times 1} \quad,\quad \overline{\boldsymbol{z}_q}_{\color{red} C \times 1} \triangleq \frac{1}{K_q} \sum_{k | y_k=q} {\boldsymbol{z}_k}_{\color{red} C \times 1} \quad,\quad K_q \triangleq \sum_{k | y_k=q} 1\,.
\end{equation}

Inspired by Linear Discriminant Analysis (LDA), we quantify separability of the classification as the ratio of between-class scatter $S_b$ and within-class scatter $S_w$.
\begin{equation}
S_b \triangleq \frac{1}{Q} \sum_q \|{\overline{\boldsymbol{z}}_q}_{\color{red} C \times 1} - \overline{\boldsymbol{z}}_{\color{red} C \times 1} \|^2 \quad,\quad S_w \triangleq \frac{1}{Q} \sum_q \frac{1}{K_q} \sum_{k | y_k=q}\| {\boldsymbol{z}_k}_{\color{red} C \times 1} - {\overline{\boldsymbol{z}}_q}_{\color{red} C \times 1} \|^2 \,.
\end{equation}

To achieve a good representation for classification, we aim to improve separability of the data points by maximizing the ratio:
\begin{equation}
\boldsymbol{p}^* \triangleq \arg\max_{\boldsymbol{p}} \frac{S_b}{S_w} \,.
\end{equation}

Plugging definitions from (\ref{eq:def_avr}) and (\ref{eq:def_z}) into the above objective function yields:
\begin{equation}
\frac{S_b}{S_w} = \frac{{\boldsymbol{p}^T}_{\color{red} 1 \times N} \sum_q \big( (\overline{\boldsymbol{X}}_q - \overline{\boldsymbol{X}})_{\color{red} N \times C} {(\overline{\boldsymbol{X}}_q - \overline{\boldsymbol{X}})^T}_{\color{red} C \times N} \big) \boldsymbol{p}_{\color{red} N \times 1}}{{\boldsymbol{p}^T}_{\color{red} 1 \times N} \big( \sum_q \frac{1}{K_q} \sum_{k | y_k=q}  ( \boldsymbol{X}_k - \overline{\boldsymbol{X}}_q)_{\color{red} N \times C} {( \boldsymbol{X}_k - \overline{\boldsymbol{X}}_q)^T}_{\color{red} C \times N} \big) \boldsymbol{p}_{\color{red} N \times 1}} \,,
\end{equation}
where $\overline{\boldsymbol{X}}, \overline{\boldsymbol{X_q}}$ are defined based on $\boldsymbol{X}_k$'s in a similar way done for $\boldsymbol{z}$ in (\ref{eq:def_avr}). For brevity, define:
\begin{eqnarray}
\boldsymbol{A}_{\color{red}N \times N} &\triangleq& \sum_q (\overline{\boldsymbol{X}}_q - \overline{\boldsymbol{X}})_{\color{red} N \times C} {(\overline{\boldsymbol{X}}_q - \overline{\boldsymbol{X}})^T}_{\color{red} C \times N} \\
\boldsymbol{B}_{\color{red}N \times N} &\triangleq& \sum_q \frac{1}{K_q} \sum_{k | y_k=q} ( \boldsymbol{X}_k - \overline{\boldsymbol{X}}_q)_{\color{red} N \times C} {( \boldsymbol{X}_k - \overline{\boldsymbol{X}}_q)^T}_{\color{red} C \times N} \,.
\end{eqnarray}

This way our goal is to maximize separability:
\begin{equation}
\boldsymbol{p}^*_{\color{red} N \times 1} =\arg\max_{\boldsymbol{p}} \frac{{\boldsymbol{p}^T}_{\color{red}1 \times N} \boldsymbol{A}_{\color{red} N \times N} {\boldsymbol{p}}_{\color{red}N \times 1}}{{\boldsymbol{p}^T}_{\color{red}1 \times N} \boldsymbol{B}_{\color{red} N \times N} {\boldsymbol{p}}_{\color{red}N \times 1}} \,.
\label{eq:sep_problem}
\end{equation}

\subsection{Closed-Form Solution}
\label{sec:closed}

The solution to (\ref{eq:sep_problem}) is ill-posed; if some  $\boldsymbol{p}^\dag$ is a solution, then so is $a \,  \boldsymbol{p}^\dag$ for any $a \neq 0$. To avoid such freedom of scale, we anchor $S_w$ to a fixed value, e.g. $1$ and then solve:
\begin{equation}
\boldsymbol{p}^* =\arg\max_{\boldsymbol{p}} {\boldsymbol{p}^T}_{\color{red}1 \times N} \boldsymbol{A}_{\color{red} N \times N} {\boldsymbol{p}}_{\color{red}N \times 1} \quad\mbox{s.t.}\quad {\boldsymbol{p}^T}_{\color{red}1 \times N} \boldsymbol{B}_{\color{red} N \times N} {\boldsymbol{p}}_{\color{red}N \times 1}=1 \,.
\end{equation}

In addition, we wish to keep $\boldsymbol{p}$ \emc{localized}  so that the operator $\boldsymbol{P}$ respects the topology of the space $\Omega$. This is important when the pooled feature maps are to be processed by convolution operators (in the future layers). We can encourage localization by introducing a penalty term of the form ${\boldsymbol{p}^T}_{\color{red} 1 \times N} \boldsymbol{C}_{\color{red} N \times N} \boldsymbol{p}_{\color{red} N \times 1}$,
where $\boldsymbol{C}$ is a diagonal matrix with positive components. How the elements of $\boldsymbol{C}$ are chosen is described in Section \ref{sec:c}.

Applying localization penalty and anchoring of $S_w$ results in the following optimization:
\begin{equation}
\boldsymbol{p}^* = \arg\max_{\boldsymbol{p}} \boldsymbol{p}^T_{\color{red}1 \times N} \boldsymbol{A}_{\color{red} N \times N} {\boldsymbol{p}}_{\color{red}N \times 1} \,\,-\, \,\rho \, \boldsymbol{p}^T \boldsymbol{C} \boldsymbol{p} \quad,\quad \mbox{s.t. } \boldsymbol{p}^T_{\color{red}1 \times N} \boldsymbol{B}_{\color{red} N \times N} {\boldsymbol{p}}_{\color{red}N \times 1}=1 \,,
\end{equation}
where $\rho>0$ is the penalty coefficient. It can be shown (see the supplementary appendix for proof) that the solution $\boldsymbol{p}^*$ must satisfy the following \emc{generalized eigenvalue problem} for some\footnote{It turns out $\alpha$ is proportional to $\rho$ and hence still serves as some penalty coefficient. See the supplementary appendix for details.} $\alpha > 0$. 
\begin{equation}
\boldsymbol{A}_{\color{red} N \times N} \boldsymbol{p}_{\color{red}N \times 1} = - \lambda \Big( \boldsymbol{B}_{\color{red} N \times N}  + \alpha \,  \boldsymbol{C}_{\color{red} N \times N} \Big) \boldsymbol{p}_{\color{red}N \times 1} \,.
\end{equation}

One way to solve the generalized eigenvalue problem is by matrix inversion. If the matrix on the r.h.s. is invertible, then we have:
\begin{equation}
\Big( \boldsymbol{B}_{\color{red} N \times N} + \alpha \,  \boldsymbol{C}_{\color{red} N \times N} \Big)^{-1} \boldsymbol{A}_{\color{red} N \times N} \boldsymbol{p}_{\color{red}N \times 1} = - \lambda \boldsymbol{p}_{\color{red}N \times 1} \,,
\end{equation}
which implies that the optimal $\boldsymbol{p}$ must be an (in fact the leading) eigevector of the following matrix:
\begin{equation}
\label{eq:final_eq}
\boldsymbol{p}^*_{\color{red}N \times 1} = \texttt{top\_eigvec\Big(} ( \boldsymbol{B}_{\color{red} N \times N}  + \alpha \, \boldsymbol{C}_{\color{red} N \times N} )^{-1} \boldsymbol{A}_{\color{red} N \times N} \texttt{\Big)} \,.
\end{equation}
Since the matrix $\boldsymbol{C}$ is diagonal with positive entries, and the matrix $\boldsymbol{B}$ is positive semi-definite, the matrix $\boldsymbol{C}$ has a regularization effect when computing the inverse of $\boldsymbol{B} + \alpha, \boldsymbol{C}$. Thus we refer to $\boldsymbol{C}$ as a regularization matrix.

\subsection{Regularization Matrix $\boldsymbol{C}$}
\label{sec:c}
We now explain how the components of $\boldsymbol{C}$ are chosen. For clarity, we temporarily (throughout this subsection) switch from the brief notation $\boldsymbol{p}$ to the full notation $\boldsymbol{p}_m$. We also need to switch from  $\boldsymbol{C}$ to $\boldsymbol{C}_m$ accordingly. Note that each component of $\boldsymbol{p}_m$ corresponds to a coordinate $(i,j)\in \Omega$. To show this relationship we use the notation $\texttt{coord}_{\Omega}(n) = (i,j)$. Similarly, for the space $\omega$, each index $m$ is associated with a coordinate $(i,j) \in \omega$, and the relationship is shown via $\texttt{coord}_{\omega}(m) = (i,j)$. We penalize the $n$'th component of $\boldsymbol{p}_m$ (recall each $\boldsymbol{p}_m$ is a vector of size $N$, thus $1 \leq n \leq N$) by its coordinate distance from that of $m$. Since in general $N \neq M$, a scale correction needs to be done. This way, the amount of penalty for the $n$'th component of $\boldsymbol{p}_m$, which is encoded in the diagonal element $c^m_{n,n}$, is set to  $c^m_{n,n} \triangleq \| \texttt{coord}_\Omega(n)\,-\, s \, \texttt{coord}_\omega(m) \|^2 $, where $s$ is the scale factor $s = \frac{I}{I^\prime}=\frac{J}{J^\prime}$. Here $c^m_{i,j}$ refers to the $(i,j)$'th component of the matrix $\boldsymbol{C}_m$.

In words, this penalty scheme means that if a point in the source feature maps contributes to a point in the destination feature map, where the latter is far from the source point, then that contribution is penalized.

\subsection{Algorithm}

The resulted procedure is shown in Algorithm \ref{algo:pooling}. Note that the matrices $\boldsymbol{A}$ and $\boldsymbol{B}$ are the same for any $m$. From the beginning of the algorithm up to line 16 is to compute these matrices. However, the matrix $\boldsymbol{C}$ and the vector $\boldsymbol{p}^*_m$ both depend on $m$ and thus Line 20 to the end loops over $m$ to compute each $\boldsymbol{C}$ and its resulted $\boldsymbol{p}^*_m$.

\begin{algorithm}
\begin{algorithmic}[1]{
\STATE{{\bf Input:} Training pairs $\cup_{k=1}^K \{(\boldsymbol{X}_k,y_k)\}$ where $\boldsymbol{X}_k\in \mathbb{R}^{N \times C}$ and $y_k \in \{1,\dots,Q\}$ with $Q$ being number of classes, penalty coefficient $\alpha>0$, scaling factor $s$.}
\STATE {$\bar{\boldsymbol{S}} \leftarrow \boldsymbol{O}_{\color{red} N \times C} \,,\, \bar{\boldsymbol{T}} \leftarrow \boldsymbol{O}_{\color{red} N \times N} \,,\, \boldsymbol{B} \leftarrow \boldsymbol{O}_{\color{red} N \times N} $}
\FOR{$q \leftarrow 1$ \TO $Q$}
\STATE {$\boldsymbol{S} \leftarrow \boldsymbol{O}_{\color{red} N \times C} \,,\, \boldsymbol{T} \leftarrow \boldsymbol{O}_{\color{red} N \times N}$}
\FOR{$k \leftarrow 1$ \TO $K_q$}
\IF{$y_k = q$}
\STATE {$\boldsymbol{S} \leftarrow \boldsymbol{S}+\boldsymbol{X}_k$}
\STATE {$\boldsymbol{T} \leftarrow \boldsymbol{T}+\boldsymbol{X}_k \boldsymbol{X}_k^T$}
\ENDIF
\ENDFOR
\STATE {$\bar{\boldsymbol{X}}_q \leftarrow \frac{1}{K_q} \boldsymbol{S}$}
\STATE {$\bar{\boldsymbol{S}} \leftarrow \bar{\boldsymbol{S}} + \bar{\boldsymbol{X}}_q$}
\STATE {$\bar{\boldsymbol{T}} \leftarrow \bar{\boldsymbol{T}} + \bar{\boldsymbol{X}}_q \bar{\boldsymbol{X}}_q^T$}
\STATE {$\boldsymbol{B} \leftarrow \boldsymbol{B} + \frac{1}{K_q} \boldsymbol{T} - \bar{\boldsymbol{X}}_q \bar{\boldsymbol{X}}_q^T$}
\ENDFOR
\STATE {$\boldsymbol{A} \leftarrow  \bar{\boldsymbol{T}} - \frac{1}{Q}\bar{\boldsymbol{S}} \bar{\boldsymbol{S}}^T$}
\STATE {{$\boldsymbol{d}_{\color{red} N \times 2} \leftarrow [\texttt{coord}_\Omega(1),\dots, \texttt{coord}_\Omega(N)]$}}
\STATE {{$\boldsymbol{e}_{\color{red} N \times 1} \leftarrow [\|\texttt{coord}_\Omega(1)\|^2,\dots, \|\texttt{coord}_\Omega(N)\|^2]$}}
\STATE {$M \leftarrow \frac{N}{s^2}$}
\FOR {$m \leftarrow 1$ \TO $M$} 
\STATE {$\boldsymbol{c}_{\color{red} N \times 1} \leftarrow \boldsymbol{e}_{\color{red} N \times 1} - 2 s   \boldsymbol{d}_{\color{red} N \times 2} {{\texttt{coord}_\omega}(m)}_{\color{red} 2 \times 1} + s^2  \|{\texttt{coord}_\omega}(m)\|^2 \boldsymbol{1}_{\color{red} N \times 1}$} 
\STATE {$\boldsymbol{C}_{\color{red} N \times N} \leftarrow \diag(\boldsymbol{c}_{\color{red} N \times 1})$}
\STATE {${\boldsymbol{p}^*_m}_{\color{red} N \times 1} \leftarrow \texttt{top\_eigvec\Big(} ( \boldsymbol{B}_{\color{red} N \times N}  + \alpha \, \boldsymbol{C}_{\color{red} N \times N} )^{-1} \boldsymbol{A}_{\color{red} N \times N} \texttt{\Big)} $}
\ENDFOR
\RETURN {$\frac{\boldsymbol{p}_1^*}{\|\boldsymbol{p}_1^*\|},\dots,\frac{\boldsymbol{p}_M^*}{\|\boldsymbol{p}_M^*\|}$}
\caption{Learning Pooling Operator.}
\label{algo:pooling}
}
\end{algorithmic}
\end{algorithm}

\subsection{Implementation Details}

{\bf Choice of Norm.} The last line of algorithm returns  $\frac{\boldsymbol{p}_1^*}{\|\boldsymbol{p}_1^*\|},\dots,\frac{\boldsymbol{p}_M^*}{\|\boldsymbol{p}_M^*\|}$. We will explore $\ell_1$ and $\ell_2$ norms in the experiments.

{\bf Normalization of Feature Maps.} The pooling operator is shared across all channels. However, the intensity values in each channel could potentially have a different center and scale, making it hard for the same pooling to provide similar effect on all channels. To fix this, we normalize feature maps before forming the matrices $\boldsymbol{X}_k$ and applying Algorithm \ref{algo:pooling}. More precisely, for a given feature map $f_k(i,j,c)$ ($k=1,\dots,K$, with $K$ being size of the training set), the normalized feature map is defined as $g_k(i,j,c) \triangleq \frac{f_k(i,j,c) - \bar{f}(c)}{\sqrt{v(c)}}$, where $\bar{f}(c) \triangleq \frac{1}{I\, J\, K} \sum_{i,j,k} f_k(i,j,c)$ and $v(c) \triangleq \frac{1}{I\, J\, K} \sum_{i,j,k} (f_k(i,j,c) - \bar{f}(c))^2$.
After the pooling operator is applied, we transform the feature map back to its
original scale and center by multiplying by $v(c)$ and adding $\bar{f}(c)$.
This helps the output of the pooling operation be consistent with rest of the network.

{\bf Use in Deep Networks}
Consider a trained deep network using some typical pooling operator. We can convert the pooling operator at any given layer to a learned one, by treating the hidden representation at that layer as the input feature maps to Algorithm \ref{algo:pooling} (after applying the channel normalization described above). We will then adapt the network weights for the learned pooling by retraining the network. This process can be repeated for multiple layers. In our experiments, however, we observe that sometimes learned pooling even at one layer can already give a boost in test accuracy.

{\bf Number of Eigenvectors.} For simplicity of presentation, Algorithm \ref{algo:pooling} uses the top eigenvector. In principle, however, top few eigenvectors could be used instead. In fact, modern architectures often double the number of output channels while down-sampling via strided convolutions. To imitate that, we chose to select the top two eigenvectors from (\ref{eq:final_eq}), which results in two feature maps per input feature map. This keeps the size of the hidden representation after pooling consistent between our
method and the common practice.

{\bf Computing the Generalized Eigenvalue.} For simpler exposition, in Section \ref{sec:closed} and also Algorithm \ref{algo:pooling} we have used matrix inversion to solve the generalized eigenvalue problem.  However, there are more efficient approaches for solving generalized eigenvalue problem without matrix inversion. In addition, we only need the top-1 or top-2 eigenvectors, which allows further efficiency in computation. There are numerical recipes that can leverage these two properties, such as \texttt{scipy.sparse.linalg.eigs} that we used for our Python implementation

{\bf Learning Pooling by SGD.}
One may wonder why not using gradient descent to optimize a total loss (sum of the usual cross-entropy and separability criterion), instead of Algorithm \ref{algo:pooling} and thus simultaneously learn network weights and pooling operator. The answer is that it is either impractical or leads to inferior performance. 
To learn the regularized pooling operator it is necessary to store the 
matrix $\boldsymbol{C}$
for each location in
the output feature map.
This would incur a memory cost of $O(M N^2)$ and would be extremely space inefficient. The performance with an un-regularized pooling map is reported in Table \ref{tab:sgd_pool} in the appendix. In performs worse than our approach in almost all cases and in some, even worse than the baseline.

\section{Experiments} 
We study the performance of our pooling operator on the SVHN \citep{svhn}
and CIFAR-10/CIFAR-100 \citep{cifar} datasets. For the SVHN dataset
we also experiment with a reduced 5\% subset to measure the
performance of our algorithm in presence of limited labelled data.
We use 2 models for our experiments, a CNN model which is a 4-layer CovnNet
and a 18 layer ResNet \citep{resnet} model. Both these models
have 3 pooling layers in which they down sample via strided convolutions.
\footnote{Additional details about the models, datasets and training can be found in the supplementary material, along with the source code.
The implementation will be open-sourced with the camera ready version.
}

\subsection{Effect on generalization}  
\label{sec:gen_effect}

Table \ref{tab:gen} shows the effect of our pooling operator on generalization.
We are able to improve on the ResNet model in all settings and with the CNN
model on both versions on the SVHN dataset. The largest gain is
observed with the CNN model on the reduced SVHN dataset of over $10\%$.
Even when the CNN model fails to improve on the CIFAR datasets, the
performance is on par with the CNN.

\begin{table}
\caption{Effect of replacing the pooling operator on generalization. We report the
mean test error and standard deviation
after averaging over 5 trials. When multiple
pooling layers are replaced, it is indicated by
separating the hyper-parameters by a comma.
Experiments
which result in improvements
are highlighted in bold.}
\centering
\begin{tabular}{llrrrrr}
\toprule
Model & Dataset & 
\text{Baseline Error} &
Pooling Layer & 
$\alpha$ &
Norm & 
Error with \\
 &&&&&& replaced pooling \\
\midrule

CNN & Reduced SVHN & 
$32.53  \pm 0.284 $
& \nth{3} & $5$ & 1 & 
$\mathbf{21.56  \pm 0.525}$\\

& SVHN & 
$10.68 \pm 0.435$ 
& \nth{2} & 25 & 2 &
$\mathbf{9.93 \pm 0.47}$
\\

& CIFAR-10 &  $14.59 \pm 0.352$  & \nth{3} & $45$ & $2$ & 
$16.08 \pm 0.191$\\

& CIFAR-100 & $45.04 \pm 0.397$  & \nth{3} & 5 & 2 & $45.13 \pm 0.483 $ \\
\midrule
ResNet & Reduced SVHN  &
 $12.38  \pm 0.33$  &
 \nth{2} & $25$ & 1 &  
 $\mathbf{10.46 \pm 0.534}$  \\
 
 & SVHN & $4.07 \pm 0.126$ &
 \nth{2} &
 $10$ & 
 $1$  &
 $ \mathbf{3.62 \pm 0.057}$ \\
 
& CIFAR-10 &
$4.57  \pm 0.123$ &
\nth{3}, \nth{1} &
$15, 65$ & $2, 1$ &
$\mathbf{4.3 \pm 0.151}$
\\
  
  & CIFAR-100 & $22.31 \pm 0.2896$ &
 \nth{2} &
 $5$ & 
 $1$  &
 $ \mathbf{21.5 \pm 0.145}$ \\
 \bottomrule
\end{tabular}
\label{tab:gen}
\vspace{-5pt}
\end{table}

\subsection{Robustness to corruptions and perturbations}
\citep{robustness} have developed a dataset of real-world corruptions to test model robustness. For these set of experiments, the model is trained on the original CIFAR-10 training set and 
evaluated on the modified test sets provided. We use the given CIFAR-10-C and CIFAR-10-P
test sets and evaluate our approach by measuring the suggested quantities. For all of these measurements, we use the original ResNet architecture as the baseline.

In Table \ref{tab:corruptions} we measure Corruption Error on the CIFAR-10-C
dataset as suggested by  \citep{robustness} .
In the bottom most row we report the average corruption for each corruption type.
We note that among others, the model with the replaced pooling operator is more robust in presence of geometric transformations for 4 out of 5 cases. We define geometric transformations as those which can move/displace pixels.

In Table \ref{tab:perturbations}, we measure how our algorithm responds to gradually
applied perturbations with the CIFAR-10-P dataset.
Each cell reports the Flip Probability, which indicates
the probability of the predicted label changing in presence of a perturbation.
The bottom row reports the Flip Rate which is the ratio of the flip probability of
our model over the flip probability of the original ResNet model. We note that
our model does better than the original network for 10 out of 14 perturbations
and for 6 out 8 geometric perturbations.

\begin{table}

\setlength{\tabcolsep}{1pt}
\caption{Measuring robustness to corruptions as defined by \citep{robustness} 
on the CIFAR-10-C test set.
Each cell lists the average test error (percentage) over 5 models for
a particular corruption and severity. We use our CIFAR-10 model as described
in Table \ref{tab:gen}. In the last row, we report the average corruption
error (CE) across all 5 severities. In this metric, lower is better and the vanilla
ResNet itself
would score $100$. We highlight corruptions for which we do better in bold.}
\centering
\begin{tabular}{lrR{20pt}R{20pt}R{20pt}R{20pt}R{20pt}R{20pt}R{20pt}R{20pt}R{20pt}R{20pt}R{20pt}R{20pt}R{20pt}R{20pt}R{20pt}}
\toprule
& &   \multicolumn{15}{c}{Corruption} \\
\cmidrule{3-17}
\scriptsize{Model}  & \scriptsize{Severity}\ \ \   &
\multicolumn{5}{c}{\small Geometric} &
\multicolumn{10}{c}{\small Non-Geometric} \\

\cmidrule(l{3pt}r{3pt}){3-7}
\cmidrule(l{3pt}r{3pt}){8-17}
&&
{\tiny Defocus.} & {\tiny Frost.} & {\tiny Motion.} & {\tiny Zoom.} & {\tiny Elastic} & {\tiny Gauss.} & {\tiny Shot.} & {\tiny Impulse.} & {\tiny Snow} & {\tiny Frost} & {\tiny Fog} & {\tiny Bright.} & {\tiny Contr.} & {\tiny Pixel.} & {\tiny Jpeg.}\\

\cmidrule{1-17}

\small{ResNet} &$1$\ \ \ &
\tiny{$4.7$} & \tiny{$39.1$} & \tiny{$8.6$} & \tiny{$11.1$} & \tiny{$8.8$} & \tiny{$19.5$} & \tiny{$12.9$} & \tiny{$14.9$} & \tiny{$9.0$} & \tiny{$8.8$} & \tiny{$4.8$} & \tiny{$4.6$} & \tiny{$5.0$} & \tiny{$6.6$} & \tiny{$12.6$}
\\
  
&$2$\ \ \ &
\tiny{$5.8$} & \tiny{$37.7$} & \tiny{$15.0$} & \tiny{$14.1$} & \tiny{$9.0$} & \tiny{$38.3$} & \tiny{$21.3$} & \tiny{$27.4$} & \tiny{$17.2$} & \tiny{$13.6$} & \tiny{$5.6$} & \tiny{$4.9$} & \tiny{$7.7$} & \tiny{$11.6$} & \tiny{$18.6$} \\

&$3$\ \ \ &
\tiny{$10.5$} & \tiny{$36.5$} & \tiny{$23.4$} & \tiny{$20.1$} & \tiny{$13.1$} & \tiny{$56.8$} & \tiny{$42.4$} & \tiny{$38.1$} & \tiny{$14.8$} & \tiny{$21.7$} & \tiny{$7.3$} & \tiny{$5.3$} & \tiny{$11.0$} & \tiny{$16.4$} & \tiny{$20.7$}
 \\

&$4$\ \ \ &
\tiny{$10.5$} & \tiny{$36.5$} & \tiny{$23.4$} & \tiny{$20.1$} & \tiny{$13.1$} & \tiny{$56.8$} & \tiny{$42.4$} & \tiny{$38.1$} & \tiny{$14.8$} & \tiny{$21.7$} & \tiny{$7.3$} & \tiny{$5.3$} & \tiny{$11.0$} & \tiny{$16.4$} & \tiny{$20.7$}
 \\
 
&$5$\ \ \ &
\tiny{$41.6$} & \tiny{$47.9$} & \tiny{$32.0$} & \tiny{$37.4$} & \tiny{$24.2$} & \tiny{$69.7$} & \tiny{$62.2$} & \tiny{$74.9$} & \tiny{$21.5$} & \tiny{$32.6$} & \tiny{$27.0$} & \tiny{$7.9$} & \tiny{$55.7$} & \tiny{$51.4$} & \tiny{$27.5$}
 \\
 
\cmidrule{1-17}

\small{ResNet} &$1$\ \ \ &
\tiny{$4.5$} & \tiny{$41.4$} & \tiny{$8.1$} & \tiny{$10.3$} & \tiny{$8.4$} & \tiny{$19.4$} & \tiny{$12.0$} & \tiny{$15.7$} & \tiny{$8.9$} & \tiny{$8.0$} & \tiny{$4.5$} & \tiny{$4.4$} & \tiny{$4.8$} & \tiny{$6.3$} & \tiny{$12.1$}
\\

\small{with} &$2$\ \ \ &
\tiny{$5.6$} & \tiny{$39.9$} & \tiny{$14.5$} & \tiny{$13.0$} & \tiny{$8.5$} & \tiny{$40.8$} & \tiny{$21.3$} & \tiny{$28.5$} & \tiny{$17.4$} & \tiny{$12.6$} & \tiny{$5.5$} & \tiny{$4.8$} & \tiny{$7.8$} & \tiny{$11.4$} & \tiny{$17.6$}
\\ 

\small{pooling} &$3$\ \ \ &
\tiny{$9.6$} & \tiny{$37.8$} & \tiny{$23.2$} & \tiny{$18.7$} & \tiny{$12.2$} & \tiny{$61.9$} & \tiny{$45.4$} & \tiny{$39.2$} & \tiny{$15.1$} & \tiny{$20.9$} & \tiny{$7.1$} & \tiny{$5.2$} & \tiny{$11.1$} & \tiny{$16.1$} & \tiny{$19.7$}
\\

\small{replaced} &$4$\ \ \ &
\tiny{$18.4$} & \tiny{$52.3$} & \tiny{$23.2$} & \tiny{$24.5$} & \tiny{$19.7$} & \tiny{$69.4$} & \tiny{$54.6$} & \tiny{$60.9$} & \tiny{$17.4$} & \tiny{$22.3$} & \tiny{$9.6$} & \tiny{$5.8$} & \tiny{$18.7$} & \tiny{$34.6$} & \tiny{$22.6$}
 \\

&$5$\ \ \ &
\tiny{$41.8$} & \tiny{$50.3$} & \tiny{$32.4$} & \tiny{$34.8$} & \tiny{$24.6$} & \tiny{$75.0$} & \tiny{$67.3$} & \tiny{$78.1$} & \tiny{$21.5$} & \tiny{$32.2$} & \tiny{$23.3$} & \tiny{$7.8$} & \tiny{$57.7$} & \tiny{$54.2$} & \tiny{$26.8$}
 \\

\cmidrule{1-17}

CE & & 
\tiny{$\mathbf{97.0}$} & \tiny{$105.1$} & \tiny{$\mathbf{99.0}$} & \tiny{$\mathbf{93.0}$} & \tiny{$\mathbf{98.0}$} & \tiny{$107.2$} & \tiny{$105.9$} & \tiny{$103.8$} & \tiny{$100.7$} & \tiny{$\mathbf{96.0}$} & \tiny{$\mathbf{91.0}$} & \tiny{$\mathbf{98.0}$} & \tiny{$102.5$} & \tiny{$103.2$} & \tiny{$\mathbf{96.0}$} \\ 
\bottomrule
\end{tabular}
\label{tab:corruptions}
\vspace{-10pt}
\end{table}

\begin{table}
\centering
\setlength{\tabcolsep}{1pt}
\caption{Measuring robustness to perturbations as defined by \citep{robustness} on
the CIFAR-10-P dataset by using our best model from Table \ref{tab:gen}.
In each cell we report Flip Probability (FP) averaged over 5 models.
In the last row we report the Flip Rate when using
default ResNet model as a baseline. When there are multiple severities we report the average. Flip probability and rate are reported out of 100, with lower being better. Improvements
are highlighted in bold.}
\begin{tabular}{lrR{20pt}R{20pt}R{20pt}R{20pt}R{20pt}R{20pt}R{20pt}R{20pt}R{20pt}R{20pt}R{20pt}R{25pt}R{25pt}R{25pt}}
\toprule
& &   \multicolumn{14}{c}{Corruption} \\
\cmidrule{3-16}
\scriptsize{Model}  & \scriptsize{Severity}\ \ \   &
\multicolumn{8}{c}{\small Geometric} &
\multicolumn{6}{c}{\small Non-Geometric} \\
\cmidrule(l{3pt}r{3pt}){3-10}
\cmidrule(l{3pt}r{3pt}){11-16}
&&
 {\tiny Scale} & {\tiny Rot.} & {\tiny Tilt} & {\tiny Tran.} & {\tiny Shear} & {\tiny Motion.} & {\tiny Zoom.} & {\tiny Ga. B} & {\tiny Bright.} & {\tiny Spatter} & {\tiny Snow} & {\tiny Shot.} & {\tiny Speckle} & {\tiny Ga. N} \\
\cmidrule{1-16}
\tiny{ResNet} &$1$\ \ \ &
\small{$3.5$} & \small{$2.92$} & \small{$0.83$} & \small{$2.1$} & \small{$1.91$} & \small{$8.43$} & \small{$0.45$} & \small{$1.26$} & \small{$0.51$} & \small{$1.56$} & \small{$2.2$} & \small{$6.86$} & \small{$6.34$} & \small{$5.01$} \\

 &$2$\ \ \ &
&&&&&&&&&&&
\small{$10.56$} & \small{$10.39$} & \small{$14.26$} \\
 
 &$3$\ \ \ &
 &&&&&&&&&&&
\small{$17.6$} & \small{$17.01$} & \small{$25.05$} \\
\cmidrule{1-16}

\tiny{ResNet}  &$1$\ \ \ &
\small{$3.25$} & \small{$2.83$} & \small{$0.76$} & \small{$2.15$} & \small{$1.91$} & \small{$8.39$} & \small{$0.38$} & \small{$1.03$} & \small{$0.5$} & \small{$1.6$} & \small{$2.29$} & \small{$6.27$} & \small{$5.82$} & \small{$4.41$} \\

 \tiny{with} &$2$\ \ \ &
&&&&&&&&&&&
\small{$9.96$} & \small{$9.84$} & \small{$13.51$}
\\
 
\tiny{pooling replaced} &$3$\ \ \ &
&&&&&&&&&&&
\small{$16.69$} & \small{$16.19$} & \small{$22.95$}
\\
\cmidrule{1-16}
\small{Flip Rate} 

& &
\tiny{$\mathbf{92.9}$} & \tiny{$\mathbf{96.9}$} & \tiny{$\mathbf{91.6}$} & \tiny{$102.4$} & \tiny{$100.0$} & \tiny{$\mathbf{99.5}$} & \tiny{$\mathbf{84.4}$} & \tiny{$\mathbf{81.7}$} & \tiny{$\mathbf{98.0}$} & \tiny{$102.6$} & \tiny{$104.1$}
&
\tiny{$\mathbf{94.0}$} & \tiny{$\mathbf{94.4}$} & \tiny{$\mathbf{92.22}$ }  \\ 
\bottomrule
\end{tabular}
\label{tab:perturbations}
\vspace{-10pt}

\end{table}

\section{Conclusion}
We propose a more general pooling operator
than currently being used in literature.
We also present an algorithm
to learn the pooling operator in closed form given the distribution of
its inputs.
Compared to pooling operations that are shared throughout
spatial dimensions, ours allows more flexibility by being spatially varying.
We replace the standard pooling operations in a CNN and a ResNet model
and see benefits in generalizations on the CIFAR-10/CIFAR-100 and SVHN datasets.
The operator is demonstrably more robust to unseen geometric
transformations, which we show by evaluating on the CIFAR-10-C and
CIFAR-10-P test sets.\\
\bibliography{main}

\begin{thebibliography}{}

\bibitem[Arandjelovic et~al., 2016]{net_vlad}
Arandjelovic, R., Gronat, P., Torii, A., Pajdla, T., and Sivic, J. (2016).
\newblock Netvlad: Cnn architecture for weakly supervised place recognition.
\newblock In {\em Proceedings of the IEEE Conference on Computer Vision and
  Pattern Recognition}, pages 5297--5307.

\bibitem[Azulay and Weiss, 2018]{weiss}
Azulay, A. and Weiss, Y. (2018).
\newblock Why do deep convolutional networks generalize so poorly to small
  image transformations?
\newblock {\em arXiv preprint arXiv:1805.12177}.

\bibitem[Boureau et~al., 2011]{ask_locals}
Boureau, Y.-L., Le~Roux, N., Bach, F., Ponce, J., and LeCun, Y. (2011).
\newblock Ask the locals: multi-way local pooling for image recognition.
\newblock In {\em ICCV'11-The 13th International Conference on Computer
  Vision}.

\bibitem[Bruna and Mallat, 2013]{bruna_mallat}
Bruna, J. and Mallat, S. (2013).
\newblock Invariant scattering convolution networks.
\newblock {\em IEEE transactions on pattern analysis and machine intelligence},
  35(8):1872--1886.

\bibitem[Cohen and Shashua, 2016]{bias}
Cohen, N. and Shashua, A. (2016).
\newblock Inductive bias of deep convolutional networks through pooling
  geometry.
\newblock {\em arXiv preprint arXiv:1605.06743}.

\bibitem[Esteves et~al., 2018a]{esteves}
Esteves, C., Allen-Blanchette, C., Zhou, X., and Daniilidis, K. (2018a).
\newblock Polar transformer networks.
\newblock In {\em International Conference on Learning Representations}.

\bibitem[Esteves et~al., 2018b]{polar}
Esteves, C., Allen-Blanchette, C., Zhou, X., and Daniilidis, K. (2018b).
\newblock Polar transformer networks.
\newblock In {\em International Conference on Learning Representations}.

\bibitem[Felzenszwalb et~al., 2009]{latent_svm}
Felzenszwalb, P.~F., Girshick, R.~B., McAllester, D., and Ramanan, D. (2009).
\newblock Object detection with discriminatively trained part-based models.
\newblock {\em IEEE transactions on pattern analysis and machine intelligence},
  32(9):1627--1645.

\bibitem[Fukushima and Miyake, 1982]{fukushima}
Fukushima, K. and Miyake, S. (1982).
\newblock Neocognitron: A self-organizing neural network model for a mechanism
  of visual pattern recognition.
\newblock In {\em Competition and cooperation in neural nets}, pages 267--285.
  Springer.

\bibitem[Girshick et~al., 2015]{dpm}
Girshick, R., Iandola, F., Darrell, T., and Malik, J. (2015).
\newblock Deformable part models are convolutional neural networks.
\newblock In {\em Proceedings of the IEEE conference on Computer Vision and
  Pattern Recognition}, pages 437--446.

\bibitem[Gong et~al., 2014]{multi_scale}
Gong, Y., Wang, L., Guo, R., and Lazebnik, S. (2014).
\newblock Multi-scale orderless pooling of deep convolutional activation
  features.
\newblock In {\em European conference on computer vision}, pages 392--407.
  Springer.

\bibitem[He et~al., 2017]{maskrcnn}
He, K., Gkioxari, G., Doll{\'a}r, P., and Girshick, R. (2017).
\newblock Mask r-cnn.
\newblock In {\em Proceedings of the IEEE international conference on computer
  vision}, pages 2961--2969.

\bibitem[He et~al., 2016a]{resnet}
He, K., Zhang, X., Ren, S., and Sun, J. (2016a).
\newblock Deep residual learning for image recognition.
\newblock In {\em Proceedings of the IEEE conference on computer vision and
  pattern recognition}, pages 770--778.

\bibitem[He et~al., 2016b]{resnetv2}
He, K., Zhang, X., Ren, S., and Sun, J. (2016b).
\newblock Identity mappings in deep residual networks.
\newblock In {\em European conference on computer vision}, pages 630--645.
  Springer.

\bibitem[Hendrycks and Dietterich, 2019]{robustness}
Hendrycks, D. and Dietterich, T. (2019).
\newblock Benchmarking neural network robustness to common corruptions and
  perturbations.
\newblock {\em arXiv preprint arXiv:1903.12261}.

\bibitem[Ioffe and Szegedy, 2015]{batchnorm}
Ioffe, S. and Szegedy, C. (2015).
\newblock Batch normalization: Accelerating deep network training by reducing
  internal covariate shift.
\newblock {\em arXiv preprint arXiv:1502.03167}.

\bibitem[J{\'e}gou et~al., 2010]{vlad}
J{\'e}gou, H., Douze, M., Schmid, C., and P{\'e}rez, P. (2010).
\newblock Aggregating local descriptors into a compact image representation.
\newblock In {\em CVPR 2010-23rd IEEE Conference on Computer Vision \& Pattern
  Recognition}, pages 3304--3311. IEEE Computer Society.

\bibitem[Kanazawa et~al., 2014]{kanazawa}
Kanazawa, A., Sharma, A., and Jacobs, D.~W. (2014).
\newblock Locally scale-invariant convolutional neural networks.
\newblock {\em CoRR}, abs/1412.5104.

\bibitem[Kingma and Ba, 2014]{adam}
Kingma, D.~P. and Ba, J. (2014).
\newblock Adam: A method for stochastic optimization.
\newblock {\em arXiv preprint arXiv:1412.6980}.

\bibitem[Krizhevsky and Hinton, 2009]{cifar}
Krizhevsky, A. and Hinton, G. (2009).
\newblock Learning multiple layers of features from tiny images.
\newblock Technical report, Citeseer.

\bibitem[Krizhevsky et~al., 2012a]{alexnet}
Krizhevsky, A., Sutskever, I., and Hinton, G.~E. (2012a).
\newblock Imagenet classification with deep convolutional neural networks.
\newblock In {\em Advances in neural information processing systems}, pages
  1097--1105.

\bibitem[Krizhevsky et~al., 2012b]{krizhevsky2012imagenet}
Krizhevsky, A., Sutskever, I., and Hinton, G.~E. (2012b).
\newblock Imagenet classification with deep convolutional neural networks.
\newblock In {\em Advances in neural information processing systems}, pages
  1097--1105.

\bibitem[Larson and Loschky, 2009]{larson2009contributions}
Larson, A.~M. and Loschky, L.~C. (2009).
\newblock The contributions of central versus peripheral vision to scene gist
  recognition.
\newblock {\em Journal of Vision}, 9(10):6--6.

\bibitem[LeCun et~al., 1998]{lenet}
LeCun, Y., Bottou, L., Bengio, Y., Haffner, P., et~al. (1998).
\newblock Gradient-based learning applied to document recognition.
\newblock {\em Proceedings of the IEEE}, 86(11):2278--2324.

\bibitem[Lee et~al., 2016]{gen_pool}
Lee, C.-Y., Gallagher, P.~W., and Tu, Z. (2016).
\newblock Generalizing pooling functions in convolutional neural networks:
  Mixed, gated, and tree.
\newblock In {\em Artificial Intelligence and Statistics}, pages 464--472.

\bibitem[Lee et~al., 2014]{aug}
Lee, C.-Y., Xie, S., Gallagher, P., Zhang, Z., and Tu, Z. (2014).
\newblock Deeply-supervised nets.
\newblock {\em arXiv preprint arXiv:1409.5185}.

\bibitem[Li et~al., 2015]{beyond_spatial}
Li, C., Reiter, A., and Hager, G.~D. (2015).
\newblock Beyond spatial pooling: fine-grained representation learning in
  multiple domains.
\newblock In {\em Proceedings of the IEEE Conference on Computer Vision and
  Pattern Recognition}, pages 4913--4922.

\bibitem[Liu, 2018]{resnet_impl}
Liu, K. (2018).
\newblock pytorch-cifar.
\newblock \url{https://github.com/kuangliu/pytorch-cifar}.

\bibitem[Malinowski and Fritz, 2013]{fcn_pooling}
Malinowski, M. and Fritz, M. (2013).
\newblock Learnable pooling regions for image classification.
\newblock {\em arXiv preprint arXiv:1301.3516}.

\bibitem[Miech et~al., 2017]{learn_pool_vid}
Miech, A., Laptev, I., and Sivic, J. (2017).
\newblock Learnable pooling with context gating for video classification.
\newblock {\em arXiv preprint arXiv:1706.06905}.

\bibitem[Mobahi et~al., 2012]{hossein}
Mobahi, H., Zitnick, C.~L., and Ma, Y. (2012).
\newblock Seeing through the blur.
\newblock In {\em 2012 IEEE Conference on Computer Vision and Pattern
  Recognition}, pages 1736--1743. IEEE.

\bibitem[Netzer et~al., 2011]{svhn}
Netzer, Y., Wang, T., Coates, A., Bissacco, A., Wu, B., and Ng, A.~Y. (2011).
\newblock Reading digits in natural images with unsupervised feature learning.

\bibitem[Passalis and Tefas, 2017]{bof}
Passalis, N. and Tefas, A. (2017).
\newblock Learning bag-of-features pooling for deep convolutional neural
  networks.
\newblock In {\em Proceedings of the IEEE International Conference on Computer
  Vision}, pages 5755--5763.

\bibitem[Saeedan et~al., 2018]{dpp}
Saeedan, F., Weber, N., Goesele, M., and Roth, S. (2018).
\newblock Detail-preserving pooling in deep networks.
\newblock In {\em Proceedings of the IEEE Conference on Computer Vision and
  Pattern Recognition}, pages 9108--9116.

\bibitem[Sifre and Mallat, 2013]{sifre_mallat}
Sifre, L. and Mallat, S. (2013).
\newblock Rotation, scaling and deformation invariant scattering for texture
  discrimination.
\newblock In {\em Proceedings of the IEEE conference on computer vision and
  pattern recognition}, pages 1233--1240.

\bibitem[Sun et~al., 2017]{leap}
Sun, M., Song, Z., Jiang, X., Pan, J., and Pang, Y. (2017).
\newblock Learning pooling for convolutional neural network.
\newblock {\em Neurocomputing}, 224:96--104.

\bibitem[Worrall et~al., 2017]{worrall}
Worrall, D.~E., Garbin, S.~J., Turmukhambetov, D., and Brostow, G.~J. (2017).
\newblock Harmonic networks: Deep translation and rotation equivariance.
\newblock In {\em Proceedings of the IEEE Conference on Computer Vision and
  Pattern Recognition}, pages 5028--5037.

\bibitem[Zhang, 2019]{zhang}
Zhang, R. (2019).
\newblock Making convolutional networks shift-invariant again.
\newblock {\em arXiv preprint arXiv:1904.11486}.

\bibitem[Zhou et~al., 2017]{vid_bof}
Zhou, Y., Sun, X., Liu, D., Zha, Z., and Zeng, W. (2017).
\newblock Adaptive pooling in multi-instance learning for web video annotation.
\newblock In {\em Proceedings of the IEEE International Conference on Computer
  Vision}, pages 318--327.

\end{thebibliography}
\bibliographystyle{apalike}

\newpage
\section{Supplementary Appendix}
\subsection{Derivation of Closed Form}

The goal is to solve the following optimization:
\begin{equation}
\boldsymbol{p}^* = \arg\max_{\boldsymbol{p}} \boldsymbol{p}^T_{\color{red}1 \times N} \boldsymbol{A}_{\color{red} N \times N} {\boldsymbol{p}}_{\color{red}N \times 1} \,\,-\, \,\rho \boldsymbol{p}^T \boldsymbol{C} \boldsymbol{p} \quad,\quad \mbox{s.t. } \boldsymbol{p}^T_{\color{red}1 \times N} \boldsymbol{B}_{\color{red} N \times N} {\boldsymbol{p}}_{\color{red}N \times 1}=1 \,,
\end{equation}

Using Lagrange multiplier $\lambda$, the optimization has the following Lagrangian:
\begin{equation}
L \,=\, \boldsymbol{p}^T_{\color{red}1 \times N} \boldsymbol{A}_{\color{red} N \times N} \boldsymbol{p}_{\color{red}N \times 1} \,+\, \lambda (\boldsymbol{p}^T_{\color{red}1 \times N} \boldsymbol{B}_{\color{red} N \times N} \boldsymbol{p}_{\color{red}N \times 1} - 1) \,-\, \rho \, \boldsymbol{p}^T_{\color{red}1 \times N} \boldsymbol{C}_{\color{red} N \times N} \boldsymbol{p}_{\color{red}N \times 1} \,.
\end{equation}

The derivative of $L$ w.r.t. $\boldsymbol{p}$ is:
\begin{equation}
\frac{\partial L}{\partial \boldsymbol{p}} = \boldsymbol{A}_{\color{red} N \times N} \boldsymbol{p}_{\color{red}N \times 1} + \lambda \Big( \boldsymbol{B}_{\color{red} N \times N}  - \frac{\rho}{\lambda} \, \boldsymbol{C}_{\color{red} N \times N} \Big) \boldsymbol{p}_{\color{red}N \times 1} \,.
\end{equation}

It is not difficult to verify that $\lambda \leq 0$. Hence, defining $\alpha \triangleq - \frac{\rho}{\lambda}$, we learn that $\alpha \geq 0$ (because $\rho > 0)$.
\begin{equation}
\frac{\partial L}{\partial \boldsymbol{p}} = \boldsymbol{A}_{\color{red} N \times N} \boldsymbol{p}_{\color{red} N \times 1} + \lambda \Big( \boldsymbol{B}_{\color{red} N \times N} + \alpha \,  \boldsymbol{C}_{\color{red} N \times N} \Big) \boldsymbol{p}_{\color{red}N \times 1} \,.
\end{equation}

\subsection{Dataset details}
 In our descriptions,
an epoch indicates the number of steps necessary to perform one full pass over the training
data. Whenever reduced datasets are used, the number of steps in each epoch is scaled 
accordingly. To choose  hyper-parameters we use cross validation performed
by evaluating performance on 5 distinct random held-out subsets of the training data. 
\begin{itemize}
  \item {\bf CIFAR-10/CIFAR-100} 
For both the CIFAR datasets, we normalize
the images by subtracting the mean and dividing by the standard deviation over the entire
training set.
During training, images are augmented  using the technique described in \citep{aug} which 
consists of padding images by 4 pixels and randomly cropping a $32 \times 32$ piece along with 
adding horizontal flips. All models on the CIFAR datasets are trained for $350$ epochs
with the learning for the ResNet model decayed at epochs $150$ and $250$. 
For cross-validation
we use 10 \% subsets containing 5000 samples.
\item {\bf SVHN} For the SVHN dataset each image is normalized separately with no additional
data-augmentation applied.
All models on this dataset are trained for
$150$ epochs with the learning rate for the ResNet model decayed at epochs $50$ and $100$.
Cross validation is done by holding out 5 \% of the training data.
\item {\bf Reduced SVHN} We use this dataset to measure the performance of our algorithm
in presence of less labelled samples.
This is a reduced version of the SVHN dataset in which we only train
with 5 \% of the training data. For cross validation, 20 \% of the reduced dataset is held out.
\end{itemize}

\subsection{Model Details}
We use our pooling operator within two models. These models perform spatial pooling by using
stride 2 convolutions, and we experiment with replacing the 3 different layers in which they reduce spatial dimensions, with the exception of the final average pooling layer: 

\begin{itemize}
\item  {\bf CNN} This is a 4-layer ConvNet with convolution kernels of  size $3 \times 3$.
The first convolution uses $64$ channels and is followed by a ReLU non-linearity.
The 3 convolution layers which follow are strided convolutions with a stride of 2 (to reduce spatial dimensions) and double the number of channels from the previous layer, each of
them followed by batch-norm \citep{batchnorm} and ReLU. Towards the end, the feature map
is aggregated via global average pooling and fed into a linear layer which outputs logits.
The CNN model is trained using the Adam optimizer \citep{adam} with a fixed learning rate of
$0.001$.
\item {\bf ResNet} The second architecture we use is a 18 layer ResNet described \citep{resnetv2} with its hyper parameters chosen form the implementation by \citep{resnet_impl}.  The network is trained with SGD and momentum coefficient of $0.9$
and a starting learning rate of $0.1$, decayed a factor of $10$
after fixed number of epochs for each dataset.
\end{itemize}

\subsection{Training Procedure}
\label{sec:training}
As the first step of our algorithm we train our models till convergence using the default
pooling operation in each model. This is
followed by using Algorithm 1 (in main paper) to compute
our pooling operator along with the normalization parameters $f$ and $g$.
While estimating matrices $\boldsymbol{A}$ and $\boldsymbol{B}$, for
both models, we use at most $10000$ samples per class.
We then replace each pooling layer, one at a time, with our own pooling operator with
various values of $\alpha$ and choice of norm, and re-train the network from a random initialization.
We choose the setting that leads to the best average cross-validation error. Using this setting, we train on the full training dataset
and report numbers on the test set.

It is possible to use this procedure multiple times to replace more than one pooling layer. 
In our experiments, we tried replacing multiple pooling layers while using the CIFAR-10 and
CIFAR-100 datasets. Only on CIFAR-10, replacing with \nth{3} and the \nth{1} pooling layer
respectively in a ResNet led to a non-trivial reduction in cross-validation error. 
For all other models and datasets we report performance after replacing only a single pooling layer
inside the model.

\newpage
\subsection{Additional visualizations}
\subsubsection{ResNet on reduced SVHN}
\begin{figure}[h]
\centering
\includegraphics[width=0.5\textwidth]{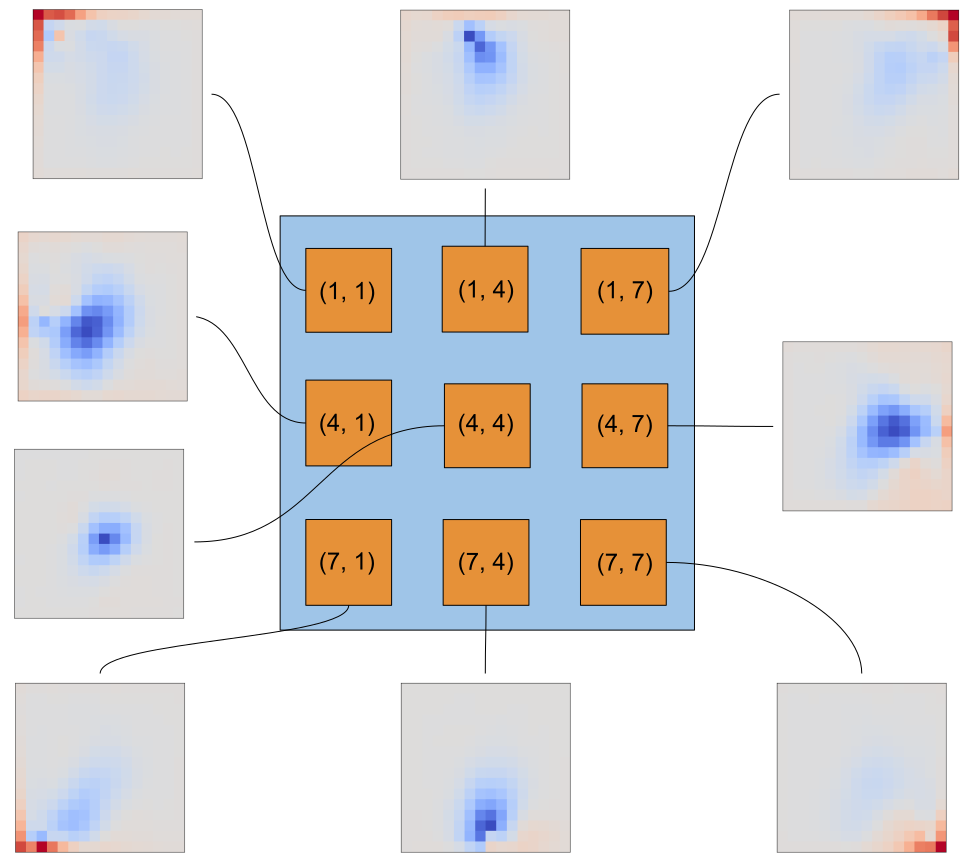}
\caption{Visualization of best performing pooling map on ResNet mode
with Reduced SVHN dataset.}
\end{figure}

\subsubsection{ResNet on CIFAR-100}
\begin{figure}[h]
\centering
\includegraphics[width=0.5\textwidth]{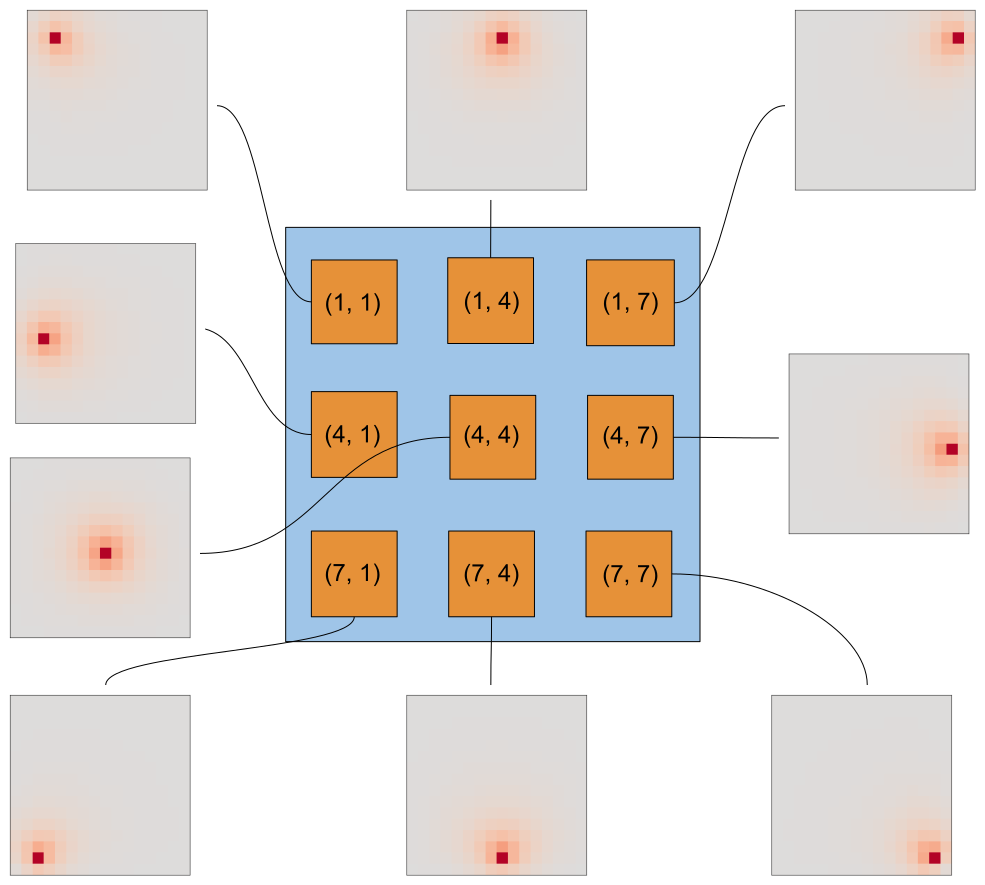}
\caption{Visualization of best performing pooling map on ResNet with the
CIFAR-100 dataset.}
\end{figure}

\subsection{Comparing operator learned
through SGD}

\begin{table}
\caption{Effect of learning a pooling map through SGD.
To keep the results comparable with Table \ref{tab:gen} we learned 2 distinct pooling maps
which has the effect of doubling the number of input
channels while down-sampling.
We chose the pooling
layer by using cross-validation as described in Section 
\ref{sec:training}
}
\centering
\begin{tabular}{llrrr}
\toprule
Model & Dataset & 
\text{Baseline Error} &
Pooling Layer & 
Error with \\
 &&&& SGD \\
\midrule

CNN & Reduced SVHN & 
$32.53  \pm 0.284 $ 
& \nth{3}  & 
$22.49 \pm 0.49$
\\

& SVHN & 
$10.68 \pm 0.435$ &
\nth{3} & $ 9.19 \pm 0.218 $ 
\\

& CIFAR-10 &  $14.59 \pm 0.352$  & \nth{3} &
$16.77 \pm 0.796$
\\

& CIFAR-100 & $45.04 \pm 0.397$  & \nth{3} &
$48.17 \pm 0.363$
\\

\midrule
ResNet & Reduced SVHN  &
 $12.38  \pm 0.33$  &
 \nth{3} & 
 $ 13.78 \pm 0.357  $
 \\
 
 & SVHN & $4.07 \pm 0.126$ & \nth{3} & 
 $3.97  \pm 0.09$
\\
 
& CIFAR-10 &
$4.57  \pm 0.123$ &
\nth{1} & 
$18.33  \pm 2.612$
\\
  
  & CIFAR-100 & $22.31 \pm 0.2896$ & \nth{1} &
   $41.09  \pm 2.346$
\\
 \bottomrule
\end{tabular}
\label{tab:sgd_pool}
\vspace{-5pt}
\end{table}

\end{document}